\def\year{2020}\relax
\documentclass[letterpaper]{article} 
\usepackage{aaai20}  
\usepackage{times}  
\usepackage{helvet} 
\usepackage{courier}  
\usepackage[hyphens]{url}  
\usepackage{graphicx} 
\usepackage{graphicx}  
\usepackage{array}
\usepackage{subfigure}
\usepackage{amsmath}
\usepackage{amssymb}
\usepackage{amsthm}
\usepackage{mathtools}
\usepackage{algorithm}
\usepackage{algorithmic}
\usepackage{stfloats}
\usepackage{multirow}
\newtheorem{proposition}{Proposition}
\newcommand{\eg}{\emph{e.g. }}
\newcommand{\etal}{\emph{et.al.}}
\newcommand{\ie}{\emph{i.e. }}

\title{Distilling portable Generative Adversarial Networks for Image Translation}
\author{
		Hanting Chen$^1$, Yunhe Wang$^2$, Han Shu$^2$, Changyuan Wen$^{3}$, \\ \Large \textbf{Chunjing Xu$^2$, Boxin Shi$^{4,5}$, Chao Xu$^1$, Chang Xu$^6$}\\
		$^1$ Key Lab of Machine Perception (MOE), CMIC, School of EECS, Peking University, China,	\normalsize$^2$ Huawei Noah's Ark Lab,\\  \normalsize$^3$Huawei Consumer Business Group,
		$^4$ National Engineering Laboratory for Video Technology, Peking University,\\ \normalsize$^5$ Peng Cheng Laboratory,
		$^6$ School of Computer Science, Faculty of Engineering, The University of Sydney, Australia\\
		\small{\{chenhanting, shiboxin\}@pku.edu.cn, xuchao@cis.pku.edu.cn, c.xu@sydney.edu.au} \\
		\small{\{yunhe.wang, han.shu, wenchangyuan, xuchunjing\}@huawei.com, } \\ 
}
 
\begin{document}

\maketitle

\begin{abstract}
Despite Generative Adversarial Networks (GANs) have been widely used in various image-to-image translation tasks, they can be hardly applied on mobile devices due to their heavy computation and storage cost. Traditional network compression methods focus on visually recognition tasks, but never deal with generation tasks. Inspired by knowledge distillation, a student generator of fewer parameters is trained by inheriting the low-level and high-level information from the original heavy teacher generator. To promote the capability of student generator, we include a student discriminator to measure the distances between real images, and images generated by student and teacher generators. An adversarial learning process is therefore established to optimize student generator and student discriminator. Qualitative and quantitative analysis by conducting experiments on benchmark datasets demonstrate that the proposed method can learn portable generative models with strong performance.
\end{abstract}

\section{Introduction}

Generative Adversarial Networks (GANs) have been successfully applied to a number of image-to-image translation tasks such as image synthesis~\cite{progressivegan}, domain translation~\cite{cyclegan,pix2pix,choi2018stargan,huang2018multimodal,lee2018diverse}, image denoising~\cite{chen2018image} and image super-resolution~\cite{SRGAN}. The success of generative networks relies not only on the careful design of adversarial strategies but also on the growth of the computational capacities of neural networks. Executing most of the widely used GANs requires enormous computational resources, which limits GANs on PCs with modern GPUs. For example,~\cite{cyclegan} uses a heavy GANs model that needs about 47.19G FLOPs for high fidelity image synthesis. However, many fancy applications of GANs such as style transfer~\cite{li2016precomputed} and image enhancement~\cite{chen2018deep} are urgently required by portable devices, \eg mobile phones and cameras. Considering the limited storage and CPU performance of mainstream mobile devices, it is essential to compress and accelerate generative networks.

Tremendous efforts have been made recently to compress and speed-up heavy deep models. For example,~\cite{gong2014compressing} utilized vector quantization approach to represent similar weights as cluster centers.~\cite{wang2018learning} introduced versatile filters to replace conventional filters and achieve high speed-up ratio.~\cite{SVD} exploited low-rank decomposition to process the weight matrices of fully-connected layers.~\cite{Hash} proposed a hashing based method to encode parameters in CNNs.~\cite{wang2018packing} proposed to packing neural networks in frequency domain.~\cite{han2015deep} employed pruning, quantization and Huffman coding to obtain a compact deep CNN with lower computational complexity.~\cite{wang2017beyond} introduced circulant matrix to learn compact feature map of CNNs. ~\cite{courbariaux2016binarized,rastegari2016xnor} explored neural networks with binary weights, which drastically reduced the memory usage. Although these approaches can provide very high compression and speed-up ratios with slight degradation on performance, most of them are devoted to processing neural networks for image classification and object detection tasks.

Existing neural network compression methods cannot be straightforwardly applied to compress GANs models, because of the following major reasons.  First, compared with classification models, it is more challenging to identify redundant weights in generative networks, as the generator requires a large number of parameters to establish a high-dimensional mapping of extremely complex structures (\eg image-to-image translation~\cite{cyclegan}). Second, different from visual recognition and detection tasks which usually have ground-truth (\eg labels and bounding boxes) for the training data, GAN is a generative model that usually does not have specific ground-truth for evaluating the output images, \eg super-resolution and style transfer. Thus, conventional methods cannot easily excavate redundant weights or filters in GANs. Finally, GANs have a more complex framework that consists of a generator and a discriminator and the two networks are simultaneously trained following a minimax two-player game, which is fundamentally different to the training procedure of ordinary deep neural networks for classification. To this end, it is necessary to develop a specific framework for compressing and accelerating GANs.~\cite{aguinaldo2019compressing} proposed to minimize the MSE Loss between tracher and student to compress GANs, which only deal with the noise-to-image task, yet most usage of GANs in mobile devices are based on image-to-image translation task. Moreover, they do not distill knowledge to the discriminator, which takes an important part in GANs' training. 

In this paper, we proposed a novel framework for learning portable generative networks by utlizing the knowledge distillation scheme. In practice, the teacher generator is utlized for minimizing the pixel-wise and perceptual difference between images generated by student and teacher networks. The discriminator in the student GAN is then optimized by learning the relationship between true samples and generated samples from teacher and student networks. By following a minimax optimization, the student GAN can fully inhert knowledge from the teacher GAN. Extensive experiments conducted on several benchmark datasets and generative models demonstrate that generators learned by the proposed method can achieve a comparable performance with significantly lower memory usage and computational cost compared to the original heavy networks. 

\begin{figure*}[t]
	\centering
	\includegraphics[width=1.0\linewidth]{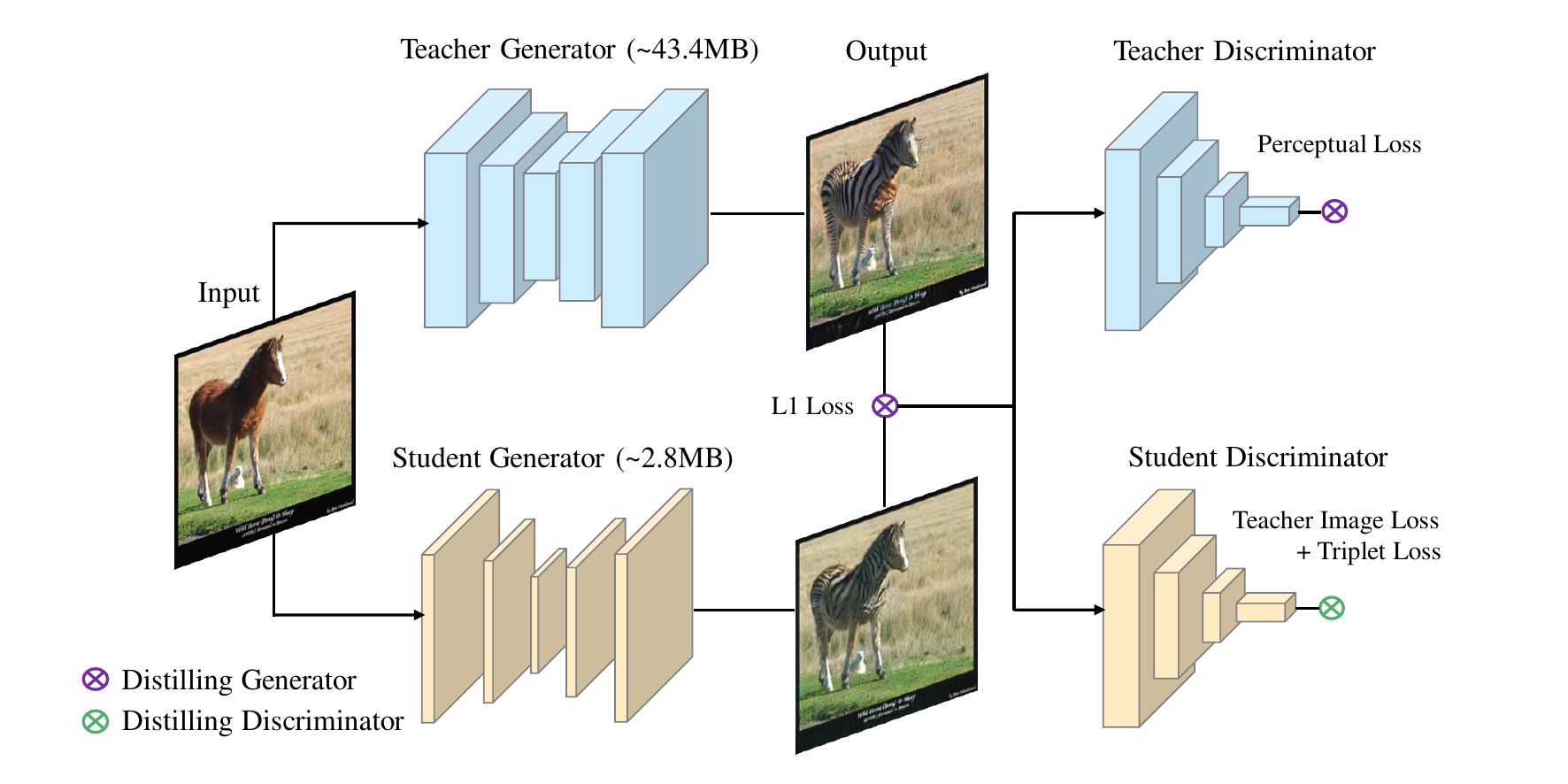}
	\caption{The diagram of the proposed framework for learning an efficient generative network by distilling knowledge from the orginal heavy network. Images generated by the student generator will be compared with those generated by the teacher generator through several metrics to fully inherit useful information from the teacher GAN.}
	\label{Fig:digram}
\end{figure*}

\section{Preliminaries}

To illustrate the proposed method, here we focus on the image-to-image translation problem and take the pix2pix~\cite{pix2pix} as an example framework. Note that the proposed algorithm does not require special component of image translation and therefore can be easily embedded to any generative adversarial networks. 

In practice, the image translation problem aims to convert an input image in the source domain $X$ to a output image in the target domain $Y$ (\eg a semantic label map to an RGB image). The goal of pix2pix is to learn mapping functions between domains $X$ and $Y$. Denote the training samples in $X$ as $\{x_1,x_2,\cdots,x_n\}$ and the corresponding samples in $Y$ as $\{y_1,y_2,\cdots,y_n\}$, the generator $G$ is optimized to maps $\mbox{x}_i$ to $\mbox{y}_i$ (\ie $G:x\rightarrow y$), which cannot be distinguished by the discriminator D. The discriminator is trained to detect the fake images generated by $G$. The objective of the GAN can be expressed as:
\begin{equation}
\begin{aligned}
\mathcal{L}_{GAN}(G,D) = &\mathbb{E}_{x,y}[\log D(x,y)]+ \\
&  \mathbb{E}_{x}[\log(1-D(x,G(x)))].
\end{aligned}
\end{equation} 
Besides fooling the discriminator, the generator is to generate images which are close to the ground truth output. Therefore, the MSE loss is introduced for $G$:
\begin{equation}
\mathcal{L}_{MSE}(G) = \mathbb{E}_{x,y}[\Vert y-G(x) \Vert_1].
\end{equation}
The entire objective of pix2pix is 
\begin{equation}
G^* =  \mbox{arg} \min_G \max_D \mathcal{L}_{GAN}(G,D) + \lambda \mathcal{L}_{MSE}(G).
\label{fcn:pix}
\end{equation}
To optimize the generator and discriminator in adversarial manner, the training of GAN is following a two-player minimax game. We alternate between optimizing $D$ with fixed $G$ and optimizing $G$ with fixed $D$. With the help of the discriminator and $L1$ loss in Fcn. (\ref{fcn:pix}), the generator can translate images from the source domain to the target domain.

Although GANs have already achieved satisfactory performance on domain translation tasks, the generators are designed to have a large number of parameters to generate images of high-dimensional semantic information, which prevents the applications of these networks in edge devices. Therefore, an effective method to learn portable GANs is urgently required.

However, GANs consisting of a generator and a discriminator, has a completely different architecture and training procedures with the vanilla CNN. It is therefore difficult to adopt existing model compression algorithms, which are developed for image recognition tasks, to handle heavy GANs model directly. Moreover, the aim of GANs is to generate images which have complex structures instead of classification or detection results. Thus, we are motivated to develop a novel framework for compressing generative models.

There are a variety of schemes for network compression such as pruning and quantization. However, these methods need special supports for achieving satisfactory compression ratio and speed improvement, which cannot be directly embedded into mobile devices. Besides eliminating redundancy in pre-trained deep models, Knowledge Distillation presents an alternative approach to learn a portable student network with comparable performance and fewer parameters by inheriting knowledge from the teacher network~\cite{hinton2015distilling,romero2014fitnets,you2017learning,wangAAAI18,heo2019knowledge}, \ie pre-trained heavy network. Therefore, we introduce the teacher-student learning paradigm (\ie knowledge distillation) to learn portable GANs with fewer parameters and FLOPs.

However, the existing teacher-student learning paradigm can only be applied to classification tasks and needs to be redesigned for the generative models which have no ground truth. Denote $G_T$ as the pretrained teacher generator and $G_S$ as the portable student generator, a straightforward method, which was proposed in ~\cite{aguinaldo2019compressing}, to adopt knowledge distillation to the student generator could be formulated as:
\begin{equation}
\label{fcn:kd}
\mathcal{L}_{L1}(G_S) = \frac{1}{n}\sum_{i=1}^n \Vert G_T(x_i)-G_S(x_i) \Vert _1^2,
\end{equation}
where $\Vert \cdot\Vert _1$ is the conventional $\ell_1$-norm. By minimizing Fcn. (\ref{fcn:kd}), images resulting from the student generator can be similar with those of the teacher generator in a pixel wise. However, this vanilla approach asking $G_S$ to minimize the Euclidean distance between the synthesis images of the teacher and student, which tend to produce blurry results~\cite{pix2pix}. This is because that the goal of Euclidean distance is to minimize all averaged plausible outputs. Moreover, GAN consists of a generator and a discriminator. Only considering the generator is not enough. Therefore, it is necessary to advance knowledge distillation to learn efficient generators. 

\section{Knowledge Distillation for GANs}

In this section, we propose a novel algorithm to obtain portable GANs utilizing the teacher-student paradigm. To transfer the useful information from the teacher GAN to the student GAN, we introduce loss functions by excavating relationship between samples and features in generators and discriminators.

\subsection{Distilling Generator}

As mentioned above, the straightforward method of utilizing the knowledge of the teacher generator is to minimize the Euclidean distance between generated images from the teacher and student generators (\ie Fcn. (\ref{fcn:kd})). However, the solutions of MSE optimization problems often lose high-frequency content, which will result in images with over-smooth textures. Instead of optimizing the pix-wise objective function,~\cite{perceptualloss} define the perceptual loss function based on the 19-th activation layer of the pertrained VGG network~\cite{vgg}. 

Motivated by this distance measure, we ask the teacher discriminator to assist the student generator to produce high-level features as the teacher generator. Compared with the VGG network which is trained for image classification, the discriminator is more relevant to the task of the generator. Therefore, we extract features of images generated by the teacher and student generators using the teacher discriminator and introduce the objective function guided by the teacher discriminator for training $G_S$:
\begin{equation}
\label{fcn:prec}
\mathcal{L}_{perc}(G_S) = \frac{1}{n}\sum_{i=1}^n \Vert \hat{D}_T(G_T(x_i))-\hat{D}_T(G_S(x_i)) \Vert _1^2,
\end{equation}
where $\hat{D}_T$ is the first several layers of the discriminator of the teacher network. Since $D_T$ has been well trained to discriminate the true and fake samples, it can capture the manifold of the target domain. The above function is more like a ``soft target'' in knowledge distillation than directly matching the generated images of the teacher and student generators and therefore is more flexible for transferring knowledge of the teacher generator. In order to learn not only low-level but also high-level information from the teacher generator, we merge the two above loss functions. Therefore, the knowledge distillation function of the proposed method for $G_S$ is 
\begin{equation}
\label{fcn:gkd}
\mathcal{L}_{KD}^G(G_S) = \mathcal{L}_{L1}(G_S) + \gamma \mathcal{L}_{perc}(G_S),
\end{equation}
where $\gamma$ is a trade-off parameter to balance the two terms of the objective.

\begin{algorithm}[t] 
	\caption{Portable GAN learning via distillation.} 
	\label{alg1} 
	\begin{algorithmic}[1] 
		\REQUIRE
		A given teacher GAN consists of a generator $G_T$ and a discriminator $D_T$, the training set $\mathcal{X}$ from domain $X$ and $\mathcal{Y}$ from domain $Y$, hyper-parameters for knowledge distillation: $\beta$ and $\gamma$. 
		\STATE Initialize the student generator $G_S$ and the student discriminator $D_S$, where the number of parameters in $G_S$ in significantly fewer than that in $G_T$;
		\REPEAT
		\STATE Randomly select a batch of paired samples $\{x_i\}_{i=1}^n$ from $\mathcal{X}$ and $\{y_i\}_{i=1}^n$ from $\mathcal{Y}$;
		\STATE Employ $G_S$ and $G_T$ on the mini-batch: \\\quad\quad\quad\quad$ z^S_i \leftarrow G_S(x_i), z^T_i \leftarrow G_T(x_i)$;	
		\STATE Employ $D_T$ and $D_S$ to compute: \\\quad$  D_S(z^S_i),D_S(z^T_i),D_S(y_i),D_T(z^S_i),D_T(z^T_i)$;
		\STATE Calculate the loss function $\mathcal{L}_{L1}(G_S)$ (Fcn. (\ref{fcn:kd})) and $\mathcal{L}_{prec}(G_S)$ (Fcn. (\ref{fcn:prec})) 
		\STATE Update weights in $G_S$ using back-propagation;
		\STATE Calculate the loss function $\mathcal{L}_{G_T}(D_S)$ (Fcn. (\ref{fcn:ds1})) and $\mathcal{L}_{tri}(D_S)$ (Fcn. (\ref{fcn:ds2})) 
		\STATE Update weights in $D_S$ according to the gradient;
		\UNTIL convergence
		\ENSURE 
		The portable generative model $G_S$.
	\end{algorithmic} 
\end{algorithm}

\subsection{Distilling Discriminator}

Besides the generator, the discriminator also plays an important role in GANs training. It is necessary to distill the student discriminator to assist training of the student generator. Different from the vanilla knowledge distillation algorithms which directly match the output of the teacher and student network, we introduce a adversarial teacher-student learning paradigm: the student discriminator is trained under the supervision of the teacher network, which will help the training of the student discriminator.
Given a well-trained GANs model, images generated by the teacher generator network can mix the spurious with the genuine. The generated images of the teacher generator $\{G(x_i)\}^n_{i=1}$ can be seen as an expansion of the target domain $Y$. Moreover, the ability of the teacher network exceeds that of the student network definitely. Therefore, images from teacher generator can be regarded as real samples for the student discriminator and the loss function for $D_S$ can be defined as:
\begin{equation}
\label{fcn:ds1}
\mathcal{L}_{G_T}(D_S) = \frac{1}{n}\sum_{i=1}^n  D_S(G_T(x_i),\mbox{\textbf{True}}).
\end{equation}

In the training of traditional GANs, the discriminator aims to classify the real images as the true samples while the fake images as the false samples, and the goal of the generator is to generate images whose outputs in the discriminator is true (\ie to generate real images). By considering images from teacher generator as real samples, Fcn. (\ref{fcn:ds1}) allows the student generator to imitate real images as well as the images generated by the teacher network, which makes the training of $G_S$ much more easier with abundant data.

As mentioned above, we regard the true images and images generated by teacher generator as the same class (\ie true labels) in $D_S$. The distance between true images and images generated by teacher generator should be smaller than that between true images and the images generated by student generator. It is natural to use triplet loss to address this problem. Triplet loss, proposed by~\cite{tripletloss}, optimizes the {black} space such that samples with the same identity are closer to each other than those with different identity. It has been widely used in various fields of computer vision such as face recognition~\cite{schroff2015facenet} and person-ReID~\cite{cheng2016person}. Therefore, we propose the triplet loss for $D_S$:
\begin{equation}
\label{fcn:ds2}
\begin{aligned}
\mathcal{L}_{tri}(D_S) = &\frac{1}{n}\sum_{i=1}^n \Big[ \Vert \hat{D}_S(y_i) - \hat{D}_S(G_T(x_i))\Vert_1 \\
-&\Vert \hat{D}_S(y_i) - \hat{D}_S(G_S(x_i))\Vert_1 + \alpha \Big]_+ ,
\end{aligned}
\end{equation}
where the $\alpha$ is the triplet margin to decide the distance between different classes, $[\cdot]_+=max(\cdot,0)$ and $\hat{D}_S$ is obtained by removing the last layer of the discriminator $D_S$. The advantage of this formulation is that the discriminator can construct a more specific manifold for the true samples than the traditional loss and then the generator will achieve higher performance with the help of the stronger discriminator.

By exploiting knowledge distillation to the student generator and discriminator, we can learn strong and efficient GANs. The overall structure of the proposed method is illstratedillustrated in Fig. (\ref{Fig:digram}). Specifically, the objective function for the student GAN can be written as follows:
\begin{equation}
\label{fcn:gs}
\begin{aligned}
\mathcal{L}_{KD}(G_S,D_S) = \mathcal{L}_{GAN}(G_S,D_S) 
+ \beta_1 \mathcal{L}_{L1}(G_S) + \\\gamma_1 \mathcal{L}_{perc}(G_S)
+\beta_2 \mathcal{L}_{G_T}(D_S) + \gamma_2 \mathcal{L}_{tri}(D_S).
\end{aligned}
\end{equation}
where the $\mathcal{L}_{GAN}$ denotes the traditional GAN loss for the generator and discriminator while $\beta_1$, $\beta_2$, $\gamma_1$ and $\gamma_2$ is the trade-off hyper-parameter to balance different objective. Note that this teacher-student learning paradigm does not require any specific architecture of GAN, and it can be easily adapted to other variants of GANs.

Following the optimization of GANs~\cite{gan}, $D_S$ and $G_S$ are trained alternatively. The objective of the proposed method is:
\begin{equation}
G_S^* =  \mbox{arg} \min_{G_S} \max_{D_S} \mathcal{L}_{KD}(G_S,D_S).
\end{equation}
By optimizing the minimax problem, the student generator can not only work cooperatively with the teacher generator but also compete adversarially with the student discriminator. In conclusion, the procedure is formally presented in Alg. (\ref{alg1}).
\begin{proposition}
	Denote the teacher generator, the student generator training with the teacher-student learning paradigm and the student generator trained without the guide of teacher as $G_T$, $G_S$ and $G'_S$, the number of parameters in $G_S$ and $G_T$ as $p_S$ and $p_T$, the number of training sample as $n$. The upper bound of the expected error of $G_S$ ($R(G_S)$) is smaller than that of $G'_S$ ($R(G'_S)$), when $n \geq \frac{p_T^4}{p_S^4}$.
	\label{prop1}
\end{proposition}

The proof of Proposition (\ref{prop1}) can be found in the supplementary materials. The inequality $ n \geq  \frac{p_T^4}{p_S^4}$ can be easily hold for deep learning whose number of training samples is large. For example, in our experiments, the number of parameters of teachers is 2 or 4 times as that of students, where $\frac{p_T^4}{p_S^4} = 16$ or $256$. The number of training samples $n$ is larger than $256$ in our experiments (\eg $n \approx3000$ in Cityscapes, $n \approx2000$ in horse to zebra task). 

\section{Experiments}
In this section, we evaluated the proposed method on several benchmark datasets with two mainstream generative models on domain translation: CycleGAN and pix2pix. To demonstrate the superiority of the proposed algorithm, we will not only show the generated images for perceptual studies but also exploit the ``FCN-score'' introduced by~\cite{pix2pix} for the quantitative evaluation. Note that~\cite{aguinaldo2019compressing} is the same as vanilla distillation in our experiments.

\begin{figure*}[t]
	\centering
	\begin{tabular}
		{>{\centering\arraybackslash}m{0.135\linewidth} >{\centering\arraybackslash}m{0.135\linewidth} >{\centering\arraybackslash}m{0.13\linewidth} >{\centering\arraybackslash}m{0.14\linewidth} >{\centering\arraybackslash}m{0.136\linewidth} >{\centering\arraybackslash}m{0.136\linewidth}}
		Input&   Ground truth&   Scratch& Aguinaldo~\etal &   Ours &  Teacher
	\end{tabular}\\
	\includegraphics[width=1\linewidth]{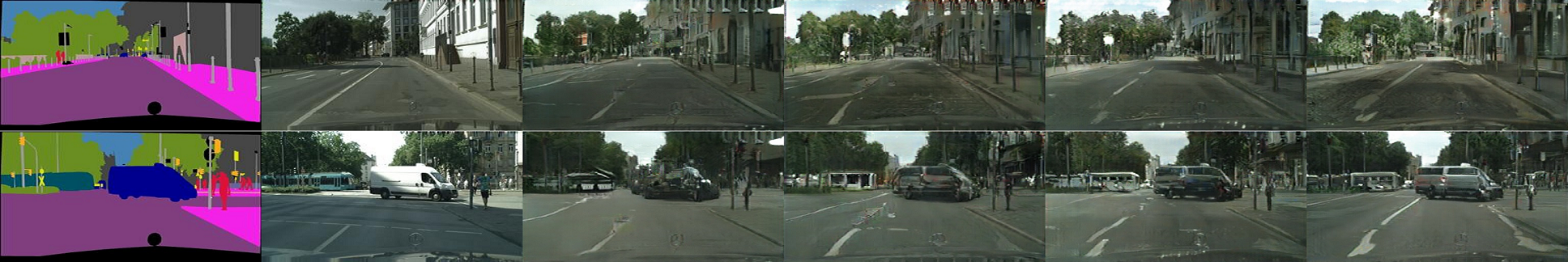}\\
	(a)Student GANs with 1/2 channels of the teacher GAN.\\
	\includegraphics[width=1\linewidth]{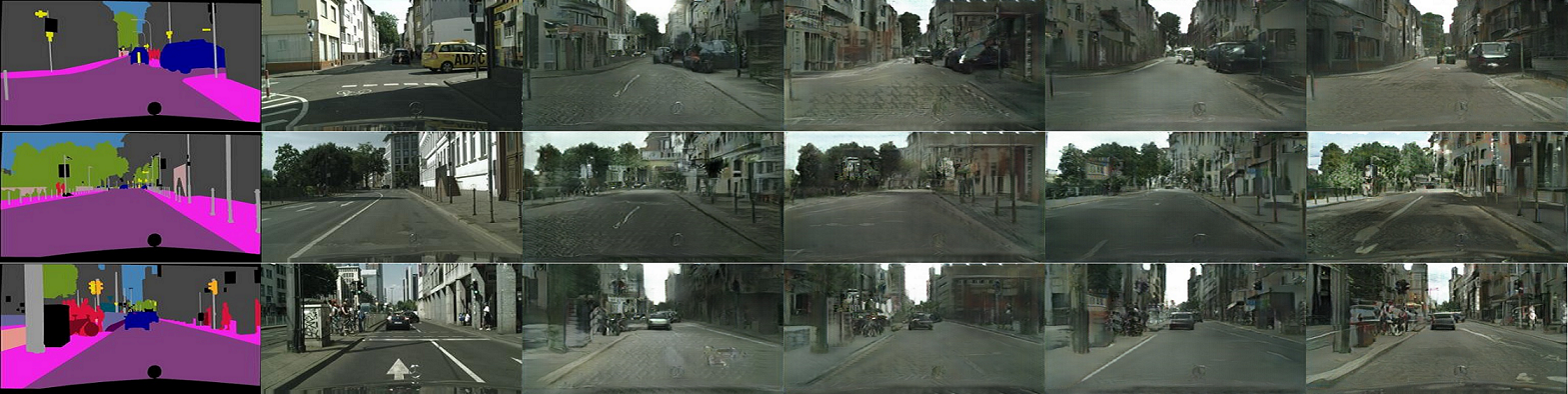}\\
	(b)Student GANs with 1/4 channels of the teacher GAN.
	\caption{Different methods for mapping labels$\rightarrow$photos trained on Cityscapes images using pix2pix.}
	\label{Fig:pix}
\end{figure*}

We first conducted the semantic label$\rightarrow$photo task on Cityscapes dataset~\cite{cityscapes} using pix2pix, which consists of street scenes from different cities with high quality pixel-level annotations. The dataset is divided into about 3,000 training images, 500 validation images and about 1,500 test images, which are all paired data. 

We followed the settings in~\cite{pix2pix} to use U-net~\cite{Unet} as the generator. The hyper-parameter $\lambda$ in Fcn. (\ref{fcn:pix}) is set to 1. For the discriminator networks, we use $70\times 70$ PatchGANs, whose goal is to classify $70\times70$ image patches instead of the whole image. When optimizing the networks, the objective value is divided by 2 while optimizing $D$. The networks are trained for 200 epochs using the Adam solver with the learning rate of 0.0002. When testing the GANs, the generator was run in the same manner as training but without dropout. 

To demonstrate the effectiveness of the proposed method, we used the U-net whose number of channels are 64 as the teacher network. We evaluated two different sizes of the student generator to have omnibearing results of the proposed method: the student generators with half channels of the teacher generator andwith $1/4$ channels. The student generator has half of the filters of the teacher. Since the discriminator is not required at inference time, we kept the structure of the student discriminator same as that of the teacher discriminator. We studied the performance of different generators: the teacher generator, the student generator trained from scratch, the student generator optimized using vanilla distillation (\ie Fcn. (\ref{fcn:kd})), and the student generator trained utilizing the proposed method.

Fig. (\ref{Fig:pix}) shows the qualitative results of these variants on the labels$\rightarrow$photos task. The teacher generator achieved satisfactory results yet required enormous parameters and computational resources. The student generator, although has fewer FLOPs and parameters, generated simple images with repeated patches, which look fake. Using vanilla distillation to minimize the $\ell_1$-norm improved the performance of the student generator, but causes blurry results. The images generated by the proposed method are much sharper and look realistic, which demonstrated that the proposed method can learn portable generative model with high quality.

\begin{table*}[t]
	\begin{center}
		\caption{FCN-scores for different methods on Cityscapes dataset using pix2pix.}
		\label{table:fcn}
		\begin{tabular}{c|c|c|c|c|c}
			\hline
			\textbf{Algorithm} &\textbf{FLOPs}&\textbf{Parameters}&\textbf{Per-pixel acc.}& \textbf{Per-class acc.}& \textbf{Class IOU}  \\
			\hline
			Teacher&$\sim$18.15G&$\sim$54.41M&52.17& 12.39& 8.20\\
			\hline
			Student from scratch&\multirow{3}{*}{$\sim$4.65G}&\multirow{3}{*}{$\sim$13.61M}&51.62& 12.10& 7.93\\
			\cline{1-1}\cline{4-6}
			~\cite{aguinaldo2019compressing}&&&50.42& 12.30& 8.00\\
			\cline{1-1}\cline{4-6}
			Student(Ours) &&&\textbf{52.22}& \textbf{12.37}& \textbf{8.11}\\
			\hline
			Student from scratch&\multirow{3}{*}{$\sim$1.22G}&\multirow{3}{*}{$\sim$3.4M}&50.80& 11.86& 7.95\\
			\cline{1-1}\cline{4-6}
			~\cite{aguinaldo2019compressing}&&&50.32& 11.98& 7.96\\
			\cline{1-1}\cline{4-6}
			Student(Ours) &&&\textbf{51.57}& \textbf{11.98}& \textbf{8.06}\\
			\hline
			Ground truth&-&-&80.42&26.72&21.13\\
			\hline
		\end{tabular}
	\end{center}
\end{table*}

\begin{figure*}
	\centering
	\begin{tabular}
		{>{\centering\arraybackslash}m{0.175\linewidth} >{\centering\arraybackslash}m{0.175\linewidth} >{\centering\arraybackslash}m{0.175\linewidth} >{\centering\arraybackslash}m{0.175\linewidth} >{\centering\arraybackslash}m{0.175\linewidth}}
		Input&     Student (Scratch)&  Aguinaldo~\etal&   Student (Ours)&  Teacher
	\end{tabular}\\
	\includegraphics[width=1\linewidth]{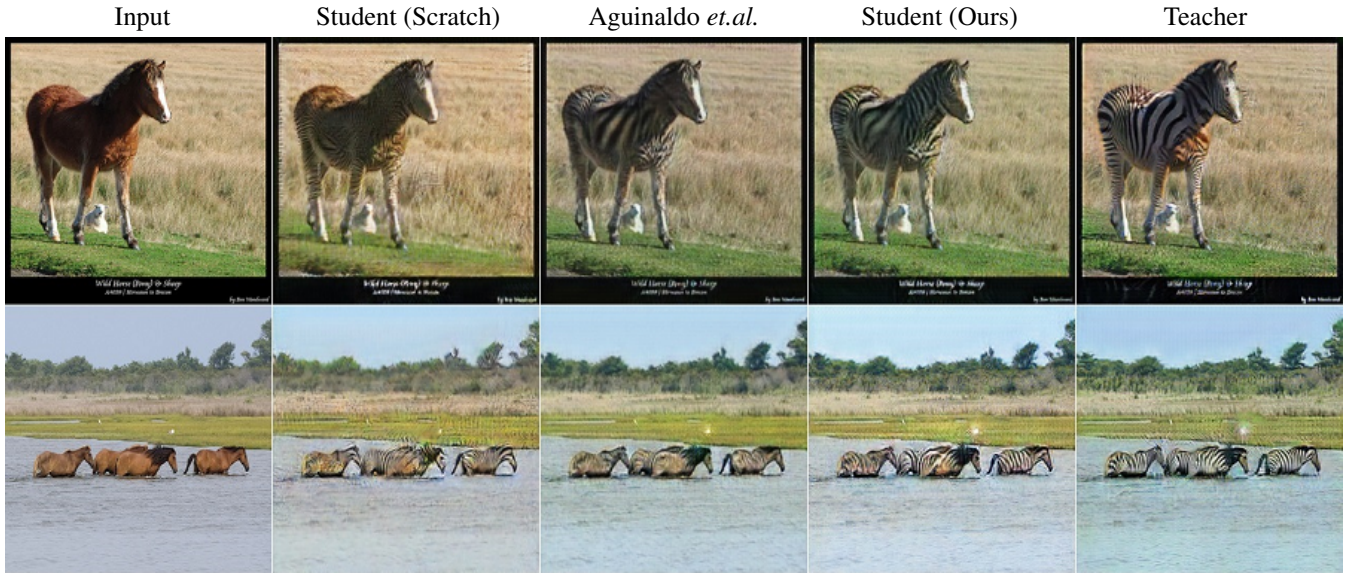}
	\caption{Different methods for mapping horse$\rightarrow$zebra trained on ImageNet images using CycleGAN.}
	\label{Fig:cycle}
\end{figure*}

\textbf{Quantitative Evaluation}
Besides the qualitative experiments, we also conducted quantitative evaluation of the proposed
method. Evaluating the quality of images generated by GANs is a difficult problem. Naive metrics such as $\ell_1$-norm error cannot evaluate the visual quality of the images. To this end, we used the metrics following~\cite{pix2pix}, \ie the ``FCN-score'', which uses a pretrained semantic segmentation model to classify the synthesized images as a pseudo metric. The intuition is that if the generated images have the same manifold structure as the true images, the segmentation model which trained on true samples would achieve comparable performance. Therefore, we adopt the pretrained FCN-8s~\cite{fcn} model on cityscapes dataset to the generated images. The results included per-pixel accuracy, per-class accuracy and mean class IOU.

Tab. (\ref{table:fcn}) reported the quantitative results of different methods. The teacher GAN achieved high performance. However, the huge FLOPs and heavy parameters of this generator prevent its application on real-world edge devices. Therefore, we conducted a portable GANs model of fewer parameters by removing half of the filters in the teacher generator. Reasonably, the student generator trained from scratch suffered degradation on all the three FCN-scores. To maintain the performance of the generator, we minimized the Euclidean distance between the images generated by the teacher network and the student network, which is shown as vanilla distillation in Tab. (\ref{table:fcn}). However, the vanilla distillation performed worse than the student generator trained from scratch, which suggests the MSE loss cannot be directly used in GAN. The proposed method utilized not only low-level but also high-level information of the teacher network and achieved a 52.22\% per-pixel accuracy, which was even higher than that of the teacher generator. 

\begin{table}[t]
	\begin{center}
		\small
		\caption{FCN-scores for different losses on Cityscapes dataset.}
		\label{table:ablation}
		\begin{tabular}{c|c|c|c}
			\hline
			\textbf{Loss}&\textbf{Per-pixel acc.}& \textbf{Per-class acc.}& \textbf{IOU}  \\
			\hline
			$\mbox{baseline}$&51.62& 12.10&7.93\\
			\hline
			$\mathcal{L}_{perc}$&51.22& 12.20& 8.01\\
			\hline
			$\mathcal{L}_{L1}+\mathcal{L}_{perc}$&51.82&12.32& 8.06\\
			\hline
			$\mathcal{L}_{G_T}$ &51.66&12.12& 8.05\\
			\hline
			$\mathcal{L}_{G_T}+\mathcal{L}_{tri}$&52.05& 12.15& 8.08\\
			\hline
			$\mathcal{L}_{L1}+\mathcal{L}_{perc}$&\multirow{2}{*}{\textbf{52.22}}& \multirow{2}{*}{\textbf{12.37}}& \multirow{2}{*}{\textbf{8.11}}\\
			$+\mathcal{L}_{G_T}+\mathcal{L}_{tri}$&&&\\
			\hline
		\end{tabular}
	\end{center}
\end{table}

\textbf{Ablation Study} We have evaluated and verified the effectiveness of the proposed method for learning portable GANs qualitatively and quantitatively. Since there are a number of components in the proposed approach, we further conducted ablation experiments for an explicit understanding. The settings are the same as the above experiments.

The loss functions of the proposed method can be divided into two parts $\mathcal{L}_{total}(G_S)$ and $\mathcal{L}_{total}(D_S)$, \ie the objective functions of the generator and the discriminator. We first evaluated the two objectives separately. As shown in Tab. (\ref{table:ablation}), the generator using $\mathcal{L}_{L1}$ loss performed better than the baseline student which was trained from scratch. By combining the perceptual loss, the student generator can learn high-level semantic information from the teacher network and achieved higher score. For the discriminator, applying the images generated from the teacher network can make the student discriminator learn a better {black} of the target domain. Moreover, the triplet loss can further improve the performance of the student GAN. Finally, by exploiting all the proposed loss functions, the student network achieved the highest score. The results of the ablation study demonstrate the effectiveness of the components in the proposed objective functions.

\textbf{Generalization Ability} In the above experiments, we have verified the performance of the proposed method on paired image-to-image translation by using pix2pix. In order to illustrate the generalization ability of the proposed algorithm, we further apply it on unpaired image-to-to image translation, which is more complex than paired translation, using CycleGAN~\cite{cyclegan}. 
We evaluate two datasets for CycleGAN: horse$\rightarrow$zebra and label$\rightarrow$photo. 

\begin{figure*}
	\centering
	\begin{tabular}
		{>{\centering\arraybackslash}m{0.175\linewidth} >{\centering\arraybackslash}m{0.175\linewidth} >{\centering\arraybackslash}m{0.175\linewidth} >{\centering\arraybackslash}m{0.175\linewidth} >{\centering\arraybackslash}m{0.175\linewidth}}
		Input&     Student (Scratch)&  Aguinaldo~\etal&   Student (Ours)&  Teacher
	\end{tabular}\\
	\includegraphics[width=1\linewidth]{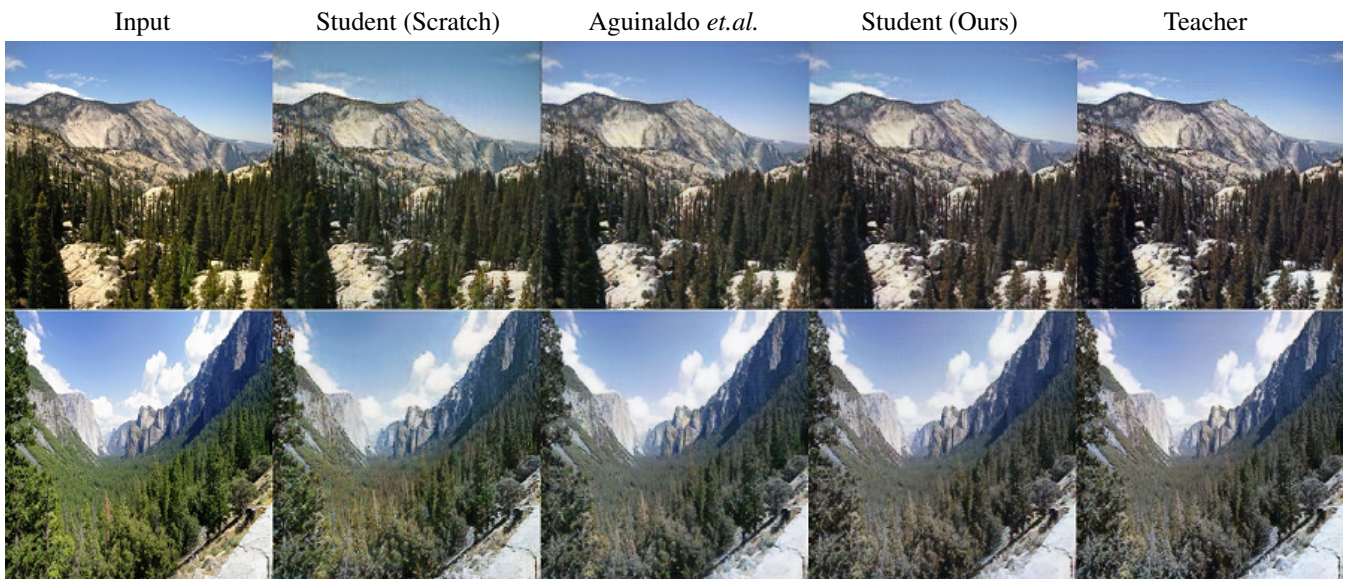}
	\caption{Different methods for mapping summer$\rightarrow$winter using CycleGAN.}
	\label{Fig:summer2winter}
\end{figure*}

For the teacher-student learning paradigm, the structure of the teacher generator was followed~\cite{cyclegan}. Note that CycleGAN has two generators to translate from domain $X$ to $Y$ and $Y$ to $X$, the number of filters of all the two student generators was set to half or quarter of that of the teacher generator. We use the same discriminator for the teacher and student network. 

Fig.~\ref{Fig:cycle} presented the images generated by different methods on the horse$\rightarrow$zebra task. Since the task is not very hard, we use an extremely portable student generators, which have only 1/4 channels of the teacher generator. The teacher generator has about 11.38M parameters and 47.19G FLOPs while the student generator has only about 715.65K parameters and 3.19G FLOPs. The images generated by the teacher network performed well while the student network trained from the scratch resulted in poor performance. The student network utilizing vanilla distillation achieved better performance, but the images were blurry. By using the proposed method, the student network learned abundant information from the teacher network and generated images better than other methods with the same architecture. The proposed method achieved comparable performance with the teacher network but with fewer parameters, which demonstrates the effectiveness of the proposed algorithm. 

We also conduct the experiments to translate summer to winter. 
The student generator trained using the proposed algorithm achieved similar performance with the teacher network but with only about 1/16 parameters. Therefore, the proposed method can learn from the teacher network effectively and generate images, which mix the spurious with the genuine, with relatively few parameters.

\section{Conclusion}

Various algorithms have achieved good performance on compressing deep neural networks, but these works ignore the adaptation in GANs. Therefore, we propose a novel framework to learning efficient generative models with significantly fewer parameters and computations by exploiting the teacher-student learning paradigm. The overall structure can be divided into two parts: knowledge distillation for the student generator and the student discriminator. We utilize the information of the teacher GAN as much as possible to help the training process of not only the student generator but also the student discriminator. Experiments on several benchmark datasets demonstrate the effectiveness of the proposed method for learning portable generative models.

\noindent\textbf{Acknowledgement}
 This work is supported by National Natural Science Foundation of China under Grant No. 61876007, 61872012 and
Australian Research Council Project DE-180101438.

{
	\bibliographystyle{aaai} \bibliography{ref}
}

\end{document}